\documentclass[11pt]{article}

\usepackage[preprint]{acl}

\usepackage{times}
\usepackage{latexsym}
\usepackage[T1]{fontenc}
\usepackage[utf8]{inputenc}
\usepackage{microtype}
\usepackage{inconsolata}
\usepackage{graphicx}
\usepackage{booktabs}
\usepackage{multirow}
\usepackage{url}
\usepackage{amsmath}
\usepackage{tikz}
\usetikzlibrary{arrows.meta, positioning, shapes.geometric, shapes.symbols,
                fit, backgrounds, calc, shadows.blur}
\usepackage{xcolor}
\usepackage{algorithm}
\usepackage{algpseudocode}
\usepackage{enumitem}
\usepackage{fontawesome5}
\usepackage{pgfplots}
\pgfplotsset{compat=1.18}
\usepgfplotslibrary{groupplots}
\usepackage{amsmath}
\usepackage{amssymb}

\title{RAILS: Retrieval-Augmented Incremental LLM Clustering at Scale}

\author{Armin Oliya \\
  Zendesk \\
  \texttt{armin.oliya@zendesk.com} \\\And
  Aleksandra Sawczuk \\
  Zendesk \\
  \texttt{ola.sawczuk@zendesk.com} \\\AND
  Radosław Białobrzeski \\
  Zendesk \\
  \texttt{radoslaw.bialobrzeski@zendesk.com} \\
  }

\begin{document}
\maketitle
\begin{abstract}
Using a Large Language Model (LLM) \emph{as} the clusterer
at production scale is hard: prompts cannot hold the entire
label space, and per-document serial processing does not
deliver the throughput real workloads require. We present
\textbf{RAILS}, a retrieval-augmented incremental LLM
clusterer that turns clustering into a simple loop over a
growing label pool and scales through document batching with
bounded concurrency. On six public benchmarks RAILS exceeds
the strongest prior LLM-clustering method on average, lifting
accuracy from $51.2\%$ to $59.3\%$, NMI from $67.2\%$ to
$74.8\%$, and ARI from $45.4\%$ to $54.7\%$. We further
report production-deployment evidence from a SaaS
ticket-topic-discovery pipeline, where RAILS
has replaced a traditional HDBSCAN stage with higher
clustering quality, transparent prompt-driven control, and
stateful incremental operation.
\end{abstract}

\begin{figure*}[t]
\centering
\resizebox{\textwidth}{!}{%
\begin{tikzpicture}[
  font=\sffamily\scriptsize,
  >=Stealth, thick,
  pool/.style   ={cylinder, shape border rotate=90, draw=teal!55, fill=teal!12,
                  aspect=0.35, minimum width=18mm, minimum height=16mm, align=center},
  batch/.style  ={rounded corners=3pt, draw=blue!55, fill=blue!10,
                  minimum width=28mm, minimum height=12mm, align=center, inner sep=2pt},
  retr/.style   ={rounded corners=4pt, draw=orange!65, fill=orange!18,
                  minimum width=28mm, minimum height=12mm, align=center, inner sep=2pt},
  llm/.style    ={rounded corners=5pt, draw=yellow!50!orange!90, fill=yellow!18,
                  minimum width=20mm, minimum height=18mm, align=center,
                  font=\sffamily\bfseries\small, inner sep=2pt},
  dec/.style    ={diamond, aspect=1.5, draw=gray!65, fill=gray!12,
                  minimum width=18mm, minimum height=12mm, align=center,
                  inner sep=0pt},
  reuse/.style  ={rounded corners=3pt, draw=green!55!black, fill=green!18,
                  minimum width=24mm, minimum height=9mm, align=center, inner sep=2pt},
  newlbl/.style ={rounded corners=3pt, draw=teal!60, fill=teal!18,
                  minimum width=24mm, minimum height=9mm, align=center, inner sep=2pt},
  noise/.style  ={rounded corners=3pt, draw=red!45, fill=red!10,
                  minimum width=24mm, minimum height=9mm, align=center, inner sep=2pt},
merge/.style ={rounded corners=4pt, draw=orange!65, fill=orange!18,
               minimum width=24mm, minimum height=11mm, align=center, inner sep=2pt},
  prompt/.style ={draw=violet!50, fill=violet!4, rounded corners=3pt,
                inner sep=2mm, align=left, font=\sffamily\scriptsize},
  trow/.style ={draw=blue!45, fill=blue!8, rounded corners=1.5pt,
              inner sep=1pt, font=\sffamily\scriptsize},
lrow/.style ={draw=teal!50, fill=teal!12, rounded corners=1.5pt,
              inner sep=1pt, font=\sffamily\scriptsize},
  conc/.style ={draw=blue!40!black!60, dashed, rounded corners=6pt,
              fill=blue!3, line width=0.7pt},
  flow/.style   ={->, line width=0.7pt, rounded corners=2pt}
]

\node[prompt] (prompt) at (7,0) {%
  \textbf{\sffamily\small LLM Prompt}\\[1pt]
  \tikz[baseline]{%
    \node[font=\sffamily\scriptsize\bfseries, anchor=north west] (h1) at (0,0)
                        {Tickets:};
    \node[trow, below=0.3mm of h1.south west, anchor=north west] (a1)
                        {\(\bullet\) \(x_1\): ``can't reset password''};
    \node[trow, below=0.3mm of a1.south west, anchor=north west] (a2)
                        {\(\bullet\) \(x_2\): ``payment didn't go through''};
    \node[trow, below=0.3mm of a2.south west, anchor=north west] (a3)
                        {\(\bullet\) \(\dots\) (\(B\) tickets)};
    \node[font=\sffamily\scriptsize\bfseries, below=1mm of a3.south west, anchor=north west] (h2)
                        {Labels:};
    \node[lrow, below=0.3mm of h2.south west, anchor=north west] (b1)
                        {\(\circ\) Login \& Password Reset};
    \node[lrow, below=0.3mm of b1.south west, anchor=north west] (b2)
                        {\(\circ\) Payment Failure};
    \node[lrow, below=0.3mm of b2.south west, anchor=north west] (b3)
                        {\(\circ\) \(\dots\) (\(C\) candidates)};
  }};
\coordinate (pUp)  at ([yshift= 8mm]prompt.west);
\coordinate (pLow) at ([yshift=-8mm]prompt.west);

\node[batch, anchor=east] (batch) at ([xshift=-18mm]pUp)  {Ticket Batch\\$\mathcal{B}=\{x_1,\dots,x_B\}$};
\node[retr,  anchor=east] (retr)  at ([xshift=-18mm]pLow) {Top-$K$ Retrieval\\\scriptsize(per ticket)};

\node[pool, anchor=east] (pool) at ([xshift=-14mm]retr.west) {Label Pool\\$\mathcal{L}$};

\node[llm, right=10mm of prompt] (llm) {\Huge\faBrain\\[5 pt]LLM};
\node[dec, right=8mm  of llm]    (dec) {Assign?};

\node[noise,  right=12mm of dec, yshift= 14mm] (noise) {Noise};
\node[reuse,  right=12mm of dec]                (reuse) {Reuse $\ell\in\mathcal{C}$};
\node[newlbl, right=12mm of dec, yshift=-14mm] (new)   {Create $\ell^{*}$};

\begin{pgfonlayer}{background}
  \node[conc, fit=(batch)(retr)(prompt), inner sep=5mm,
        label={[blue!55!black, font=\sffamily\bfseries\small]above:%
               Concurrent workers: slow-start $W\in[W_{\min},W_{\max}]$}] (cbox) {};
\end{pgfonlayer}

\node[merge, below=10mm of pool] (merge) {Merge\\(threshold $\tau$)};

\draw[flow] (pool.east) -- (retr.west);
\draw[flow] (batch.south) -- (retr.north);
\draw[flow] (batch.east) -- (pUp);
\draw[flow] (retr.east) -- node[above, font=\scriptsize]{candidates $\mathcal{C}$} (pLow);
\draw[flow] (prompt.east) -- (llm.west);
\draw[flow] (llm.east)    -- (dec.west);
\draw[flow] (dec.east) |- (noise.west);
\draw[flow] (dec.east) -- (reuse.west);
\draw[flow] (dec.east) |- (new.west);

\coordinate (busY) at (0, 0 |- merge.center);
\draw[flow] (new.south) |- (merge.east);
\draw[flow] (merge.north) -- (pool.south);

\end{tikzpicture}}%
\caption{Architecture of retrieval-augmented incremental LLM clustering.
Within each concurrency-managed worker, top-$K$ candidate labels are retrieved
from the per-brand pool $\mathcal{L}$ and packed with the ticket batch into
a single LLM prompt. The LLM (a hosted, shared service) returns one of three
outcomes per ticket: \emph{reuse} an existing candidate, \emph{create} a new
label, or mark as \emph{noise}. New labels are absorbed back into
$\mathcal{L}$ via threshold-based merging. Slow-start scheduling ramps
worker count $W$ from $W_{\min}$ to $W_{\max}$ over the run.}
\label{fig:architecture}
\end{figure*}
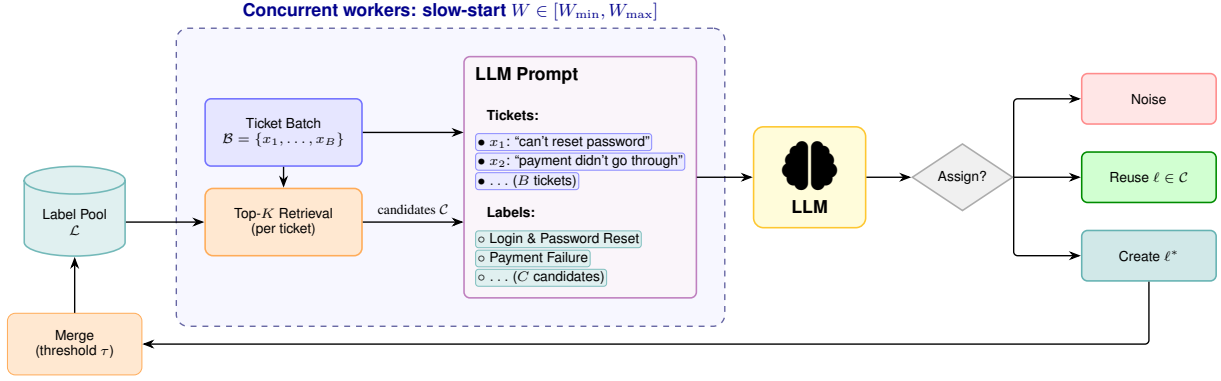

\section{Introduction}
\label{sec:intro}
Customer support platforms ingest millions of tickets per week
across thousands of customer accounts spanning industries from
e-commerce and travel to gaming and food delivery. A
\emph{topic discovery} pipeline groups these tickets into
recurring themes that downstream products consume to detect
knowledge gaps in help centres, generate and update
self-service articles, and ground generative AI copilots.
Because every downstream product inherits the topic layer's
decisions, its quality directly bounds the quality of the
entire product surface.

The conventional approach is unsupervised text clustering:
encode tickets with a pre-trained model, optionally reduce
dimensionality with UMAP~\citep{mcinnes2018umap}, and cluster
with a density- or partition-based algorithm such as
HDBSCAN~\citep{campello2013density} or $k$-means. We operated
such an embed-then-cluster pipeline in production for over a
year, across hundreds of millions of tickets, and accumulated
the structural limitations that motivate this work
(\S\ref{sec:background}).

We present \textsc{RAILS}, a \emph{retrieval-augmented
incremental LLM clustering} architecture that removes both
the embedding model and the clustering algorithm from the
critical path. An LLM directly assigns each ticket to a
growing label pool via retrieval-augmented generation. This
single design choice yields three properties the established
pipeline cannot deliver simultaneously: \textbf{native
statefulness} from the persistent label pool;
\textbf{robustness} across languages, channels (email, chat,
voice transcripts), and noisy inputs, inherited from the
LLM's pre-training; and \textbf{product-meaningful control}
via natural-language prompts that also yield a transparent
rationale for each assignment. To make this viable at
production volume, we introduce ticket batching with an
adaptive \emph{slow-start} concurrency mechanism that
reconciles label-pool quality during bootstrapping with
sustained throughput. 

We evaluate RAILS on six public clustering benchmarks where
it exceeds prior baselines,
and report production evidence from a topic-discovery
pipeline that has replaced an HDBSCAN clustering stage.

\section{Background and Problem Setting}
\label{sec:background}

The embed-then-cluster pattern scales well, but in
sustained operation, surfaces three structural limitations
that no amount of tuning could remove.

The first is a \emph{hard dependency on the encoder}: embedding
errors propagate irrecoverably to every downstream stage, and
channels the encoder was not explicitly tuned on cluster poorly(messaging
transcripts, voice call summaries, or under-resourced
languages). Any change to the embedding model
requires re-embedding and re-clustering the entire ticket history.

The second is the \emph{absence of native statefulness}. Algorithms
such as $k$-means and HDBSCAN produce a fresh partition on every run
with no notion of label persistence, so maintaining stable topic
identifiers across runs falls to post-hoc merging heuristics that are
brittle in practice.

The third is the \emph{opacity of cluster membership}: assignments
are determined by geometric proximity and a handful of
hyperparameters, not by any product-meaningful definition of a topic.
Stakeholder feedback such as ``topics are too coarse for messaging''
maps only indirectly to those hyperparameters, and answering ``why
did this ticket land here?'' requires reverse-engineering several
pipeline components. Acting on either requires time-consuming sweeps
and long feedback cycles.

\section{Related Work}
\label{sec:related}

Prior work on LLM-assisted clustering falls into three camps.

\paragraph{LLM-supervised pipelines.} ClusterLLM~\citep{clusterllm},
PRISM~\citep{prism}, and Dial-In LLM~\citep{dialinllm} use sparse LLM or
human annotations to fine-tune encoders, coherence judges, or namers that
steer a classical partitioning algorithm. These improve goal-alignment but
require a training loop that must be re-run under domain or language shift.

\paragraph{LLM-as-constraint.} A training-free line injects LLM knowledge
into an existing algorithm: group-level must/cannot-link sets for
$k$-means~\cite{constraints}, set-level similarity judgments as
features~\cite{bagoftexts}, or HDBSCAN+UMAP with multiple LLM-guided pipeline steps deployed on 90k+ chats~\cite{multiturn}. All inherit the geometric
assumptions of the underlying algorithm.

\paragraph{LLM-as-clusterer.} A third line removes the classical algorithm
entirely. GoalEx~\cite{goalex} iteratively proposes labels from corpus
subsets and classifies each document against the final set, assuming a
closed corpus. LLM-MemCluster~\cite{llmmemcluster} maintains an in-prompt
label memory with assign-or-create decisions and a separate
merge-suggestion action, switching between two prompt modes (strict vs.
relaxed) at a user-specified label ceiling $K_{\max}$. It is evaluated
serially on small benchmarks where the full memory fits in one prompt.

\textbf{RAILS} removes distance-based partitioning, fixed-$k$
constraints, and training, and exposes clustering criteria
as editable natural-language prompts. It exceeds the
strongest prior LLM methods on average across six benchmarks,
and is demonstrated to scale by an actual production
deployment in an industry setting.

\section{Method}
\label{sec:method}

Algorithm~\ref{alg:overall} and Figure~\ref{fig:architecture}
give the top-level loop. Concrete values for $B$, $K$, the
merge threshold $\tau$, and the concurrency bounds are
deployment knobs reported in \S\ref{sec:production-validation}.

\begin{algorithm}[t]
\caption{Top-level RAILS algorithm}
\label{alg:overall}
\begin{algorithmic}[1]
\Require Documents $\mathcal{D}$, label pool $\mathcal{L}$
  (initially empty or seeded), batch size $B$, retrieval
  depth $K$, concurrency bounds $W_{\min}, W_{\max}$
\State $M \gets \lceil |\mathcal{D}|/B \rceil$; split
  $\mathcal{D}$ into batches $\mathcal{B}_1, \ldots, \mathcal{B}_M$
\State assignments $\gets \emptyset$
\State $W \gets W_{\min}$; initialise semaphore $S$ with $W$ permits
\For{$j = 1, \ldots, M$ \textbf{(}up to $W$ batches in flight via $S$\textbf{)}}
  \State $\mathcal{C}_j \gets \textsc{Retrieve}(\mathcal{B}_j, \mathcal{L}, K)$
  \State $\mathbf{a}_j \gets \textsc{LLMCall}(\mathcal{B}_j, \mathcal{C}_j)$
  \Comment{one label per document}
  \State $\Lambda_j \gets \{\,a \in \mathbf{a}_j \mid a \notin
                       \mathcal{C}_j \cup \{\textsc{noise}\}\,\}$
  \Comment{newly coined}
  \State \textbf{lock}$(\mathcal{L})$:
         $(\mathcal{L}, \mathbf{a}_j) \gets
            \textsc{Merge}(\mathcal{L}, \Lambda_j, \mathbf{a}_j)$
  \State assignments $\gets$ assignments $\cup\,\mathbf{a}_j$
  \If{batch $j$ completed without provider throttling}
    \State $W \gets \min(W{+}1,\, W_{\max})$
  \EndIf
\EndFor
\State \Return assignments, final pool $\mathcal{L}$
\end{algorithmic}
\end{algorithm}

\subsection{Label pool}
\label{sec:pool}

The label pool $\mathcal{L}$ is a set of records
\[
\mathcal{L} = \big\{ (\ell_k, \, \mathbf{e}_k) \big\}_{k=1}^{|\mathcal{L}|},
\]
where $\ell_k$ is the canonical surface form of the label and $\mathbf{e}_k \in \mathbb{R}^{d}$ is its embedding, computed by the same encoder used for document embeddings (\S\ref{sec:retrieval}). The pool $\mathcal{L}$ starts either empty (cold-start) or
pre-populated with labels from a previous run or a curated
taxonomy. This is what makes RAILS streaming-friendly: successive runs over newly-arrived documents inherit prior labels via retrieval while the LLM remains free to extend the pool with genuinely new themes.

\subsection{Retrieval}
\label{sec:retrieval}

Label and document embeddings are produced by a single sentence encoder. We use it as a \emph{soft component}: its role is to surface plausible candidate labels for an LLM decision, not to determine the cluster assignment itself. A retrieval miss therefore does not produce a misclustering, only a slightly larger prompt or a duplicate-label proposal that the per-batch merge step would absorb (\S\ref{sec:merge}).

For each document $x_i \in \mathcal{B}$ with precomputed
embedding $\mathbf{q}_i$, we take its top-$K$ pool entries by
cosine similarity; the batch candidate set
$\mathcal{C}(\mathcal{B})$ is their union, of size at most
$BK$, which fits comfortably within modern LLM context
windows.

\subsection{LLM assignment and label generation}
\label{sec:llm-assignment}

A single LLM call returns one label per document
(Alg.~\ref{alg:overall}, lines 6--7); each label is either an
existing candidate, a freshly coined string, or the reserved
\textsc{noise} token. Beyond throughput, grouping $B$ documents
into a single call lets the model (i)~reuse a candidate label
across several semantically similar documents in the batch and
(ii)~coin one shared new label rather than $B$ near-duplicates.

The system prompt encodes the clustering task and granularity
target -- with cross-domain examples at the desired specificity
level -- an explicit definition of \textsc{noise} as
content-free inputs (auto-generated stubs, gibberish), and the
output JSON schema. The user prompt contains the batch and the
candidate set.

\subsection{Per-batch label merge}
\label{sec:merge}
Even with the candidate set $\mathcal{C}(\mathcal{B})$ in
the prompt, the LLM can coin a synonym of a label that did
not surface in top-$K$, and concurrent batches can
independently coin synonyms for the same emerging topic; an
embedding-based merge is the safety net that keeps the pool
from fragmenting.

For every new label $\ell \in \Lambda$, let $\ell_{k^*}$ be
its nearest pool entry by cosine similarity. If
$s(\mathbf{e}(\ell), \mathbf{e}_{k^*}) > \tau$, $\ell$ is
\emph{absorbed} into $\ell_{k^*}$: all documents in the
batch assigned to $\ell$ are reassigned to $\ell_{k^*}$ and
$\ell$ is discarded.

The merge runs under a write lock on $\mathcal{L}$ and is the
only operation that mutates the pool; it always sees the
latest committed state, so synonyms produced by concurrent
batches are caught by whichever batch commits second.

\subsection{Concurrency}
\label{sec:concurrency}
Next to document batching, concurrency is the main lever for
fitting RAILS to production throughput requirements. We gate
batch execution with a semaphore that starts at $W_{\min}$
permits (in practice $W_{\min}{=}1$) and grows by one on each
throttle-free completion up to $W_{\max}$. The slow start
serves two purposes. First, it manages cold-start label-pool
quality: the near-serial first batches populate $\mathcal{L}$
with the most common topics; once those are present, later
batches retrieve them as candidates and the LLM defaults to
reuse, instead of many in-flight workers independently coining
synonyms for the same emerging topic against a near-empty
pool. Second, it is a graceful way to respect provider rate
limits or self-hosted GPU capacity by ramping up only as fast
as the backend tolerates. More
sophisticated controllers are possible; we report this simple
mechanism as sufficient for our deployment.

\section{Experimental Setup}
\label{sec:setup}

\begin{table*}[t]
\centering
\resizebox{\textwidth}{!}{
\begin{tabular}{l|ccc|ccc|ccc|ccc|ccc|ccc|ccc}
\toprule
\multirow{2}{*}{Method} & \multicolumn{3}{c|}{ArxivS2S} & \multicolumn{3}{c|}{Massive-I} & \multicolumn{3}{c|}{MTOP-I} & \multicolumn{3}{c|}{Massive-D} & \multicolumn{3}{c|}{FewNerd} & \multicolumn{3}{c|}{FewRel} & \multicolumn{3}{c}{AVG} \\
\cmidrule(lr){2-4} \cmidrule(lr){5-7} \cmidrule(lr){8-10} \cmidrule(lr){11-13} \cmidrule(lr){14-16} \cmidrule(lr){17-19} \cmidrule(lr){20-22}
& ACC & NMI & ARI & ACC & NMI & ARI & ACC & NMI & ARI & ACC & NMI & ARI & ACC & NMI & ARI & ACC & NMI & ARI & ACC & NMI & ARI \\
\midrule
K-Means & 6.5 & 24.7 & 0.1 & 55.2 & 71.7 & 39.8 & 33.8 & 70.3 & 25.0 & 58.8 & 63.9 & 39.9 & 27.2 & 41.9 & 5.9 & 35.4 & 52.0 & 22.5 & 36.2 & 54.1 & 22.2 \\
HDBSCAN & 3.2 & 11.5 & 0.0 & 28.1 & 48.8 & 3.0 & 36.5 & 46.7 & 7.4 & 32.1 & 45.3 & 5.2 & 25.9 & 7.8 & 1.6 & 15.4 & 31.4 & 0.7 & 23.5 & 31.9 & 3.0 \\
BERTopic & 4.8 & 17.7 & 0.0 & 50.8 & 68.5 & 24.5 & 40.0 & 66.9 & 22.6 & 32.4 & 60.0 & 18.3 & 35.4 & 40.1 & 10.2 & 30.4 & 47.2 & 6.4 & 32.3 & 50.1 & 13.7 \\
ClusterLLM & 25.1 & 50.5 & 13.7 & 55.5 & \underline{74.6} & 43.2 & 36.0 & 73.4 & 30.0 & 52.4 & 65.3 & 40.8 & 37.3 & 53.1 & 10.7 & \underline{43.8} & 59.6 & 30.4 & 41.7 & 62.8 & 28.1 \\
MemCluster & 28.4 & 57.4 & 16.3 & 54.8 & 73.5 & \underline{47.9} & 64.0 & 77.5 & 68.9 & 57.6 & 67.7 & 53.8 & 59.3 & 63.3 & 53.1 & 43.2 & 63.6 & \underline{32.7} & 51.2 & 67.2 & 45.4 \\
\midrule
RAILS (Gemma 4 26B MoE) & \textbf{33.1} & \underline{58.4} & \underline{19.7} & 54.4 & 70.8 & 44.8 & \underline{71.7} & 80.2 & 76.0 & 62.5 & 68.3 & 55.2 & \textbf{65.1} & \underline{71.5} & \textbf{66.3} & 41.5 & \underline{65.2} & 27.0 & \underline{54.7} & 69.1 & \underline{48.2} \\
RAILS (Nova Lite) & 23.8 & 50.4 & 11.8 & 47.8 & 66.8 & 38.2 & 64.9 & 76.4 & 66.2 & 65.5 & 67.6 & 56.5 & 58.4 & 64.1 & 55.9 & 32.1 & 59.2 & 20.3 & 48.8 & 64.1 & 41.5 \\
RAILS (Haiku 4.5) & 28.3 & 56.2 & 15.1 & \underline{56.2} & 72.6 & 45.9 & 71.2 & \underline{81.3} & \underline{78.5} & \underline{66.4} & \underline{70.3} & \underline{56.6} & \underline{63.8} & 70.9 & \textbf{66.3} & 38.6 & 64.1 & 26.7 & 54.1 & \underline{69.2} & \underline{48.2} \\
RAILS (Sonnet 4.5) & \underline{32.5} & \textbf{62.9} & \textbf{20.1} & \textbf{60.4} & \textbf{77.5} & \textbf{54.3} & \textbf{76.2} & \textbf{85.1} & \textbf{84.1} & \textbf{70.6} & \textbf{76.4} & \textbf{60.8} & 60.0 & \textbf{72.7} & \underline{62.5} & \textbf{56.1} & \textbf{74.3} & \textbf{46.2} & \textbf{59.3}
& \textbf{74.8} & \textbf{54.7} \\
\bottomrule
\end{tabular}
}
\caption{Clustering performance on six benchmark datasets. Best results in \textbf{bold}, second best \underline{underlined}. RAILS results are mean over 3 independent runs.}
\label{tab:public_results}
\end{table*}

We evaluate RAILS on six public clustering benchmarks.
Unless otherwise noted, all runs use batch size $B{=}15$,
top-$K{=}5$, merge threshold $\tau{=}0.85$, temperature~0
for deterministic decoding, a cold-start pool
$\mathcal{L}^{(0)}{=}\emptyset$, and slow-start concurrency
bounds $W_{\min}{=}1$, $W_{\max}{=}15$. Retrieval and merge
share a single sentence encoder, \texttt{all-mpnet-base-v2}%
\footnote{\url{https://huggingface.co/sentence-transformers/all-mpnet-base-v2}}.
For each dataset we author a per-benchmark prompt that fixes
the task framing (e.g.\ \emph{paper subject area},
\emph{utterance intent}, \emph{relation type}) and provides
granularity examples tailored to its label space, while
holding the noise definition and JSON schema fixed across
all six -- the protocol used by prior LLM-clustering
work~\citep{llmmemcluster}; full prompts in
Appendix~\ref{app:public_prompts}.

We report RAILS results in Table~\ref{tab:public_results} as
the mean over three runs for each of four LLMs spanning
families and price points: Amazon Nova Lite~\cite{amazon2024nova},
Claude Haiku~4.5~\cite{antropic2025haiku}, Claude Sonnet~4.5~\cite{anthropic2025sonnet}, and Gemma~4 26B MoE~\cite{google2026gemma4}
(self-hosted via vLLM~\cite{vllm}).

We compare against standard clustering baselines and recent
LLM-based methods. \textbf{K-Means} is given the true number of
clusters $k$. \textbf{HDBSCAN} is tuned via grid search over
\texttt{min\_cluster\_size}$\in\{5,10,15,20,30,50\}$ and
\texttt{min\_samples}$\in\{1,5,10\}$, selecting the configuration
that maximises NMI. \textbf{BERTopic} uses its default UMAP and
HDBSCAN settings with automatic cluster detection. All three
consume the same pre-computed MPNet embeddings as RAILS. We also
compare against \textbf{ClusterLLM}~\citep{clusterllm} and
\textbf{MemCluster}~\citep{llmmemcluster}, reporting their
published numbers on the same datasets.

\paragraph{Metrics.} Following~\citet{clusterllm}, we report
clustering accuracy (ACC) under Hungarian matching between predicted
and gold labels, normalised mutual information (NMI), and adjusted
rand index (ARI). RAILS produces a variable label count per run;
matching is computed against the labels that survive the merge step,
with unmatched predicted labels counted as errors.
\subsection{Datasets}
\label{sec:datasets}
We evaluate on six public benchmarks spanning scientific
literature (\texttt{arxiv}~\citep{muennighoff2023arxiv}),
intent classification
(\texttt{MTOP-I}~\citep{li2021mtop},
\texttt{Massive-I} and \texttt{Massive-D}~\citep{fitzgerald2023massive}),
and entity/relation extraction
(\texttt{few\_nerd}~\citep{ding2021fewnerd},
\texttt{few\_rel}~\citep{han2018fewrel}),
with 18--102 clusters and 2{,}974--4{,}480 samples per dataset.
We use the processed splits released by~\citet{clusterllm} to
enable direct comparison with prior LLM-clustering
work~\citep{llmmemcluster}; per-dataset statistics in
Appendix~\ref{app:datasets}.

\section{Results}
\label{sec:results}

\subsection{Public Benchmarks}
\label{sec:results-public}

Table~\ref{tab:public_results} shows that RAILS improves on most metrics across all six datasets, beating both prior LLM-based methods (ClusterLLM, MemCluster) and traditional baselines. With Sonnet 4.5, RAILS reaches 59.3\% ACC / 74.8\% NMI / 54.7\% ARI on average, a +8.1-point ACC gain over the previous best (MemCluster). Gains are largest on conversational domain and fine-grained semantic tasks: Massive-D improves from 57.6\% to 70.6\% ACC and FewRel relation classification improves from 43.2\% to 56.1\% ACC, where stronger LLM reasoning matters most for disambiguating subtle distinctions.
\subsection{LLM and Embedder Choice}
\label{sec:llm-embedder}

RAILS quality scales smoothly with the LLM backbone:
Sonnet~4.5, Haiku~4.5, Gemma~4 26B MoE (self-hosted), and
Nova~Lite reach $74.8\%$, $69.2\%$, $69.1\%$, and $64.1\%$
average NMI respectively (Table~\ref{tab:public_results}),
with the gap widening on FewRel (NMI $74.3 / 64.1 / 65.2 /
59.2$). Gemma matching Haiku shows that the method is also
competitive with self-hosted open-weight models. The
clustering algorithm is unchanged across backbones, so
operators can pick along the quality--cost frontier that
fits their volume and SLA (cost estimates in
Appendix~\ref{app:cost}).

RAILS is robust along two axes that matter for deployment.
Across three encoders -- \texttt{all-mpnet-base-v2},
Instructor, and an internal intent-tuned encoder (\S\ref{sec:production-validation}) --
dataset-average NMI varies by under one point on every LLM
(e.g.\ $74.4$--$74.8$ on Sonnet, $63.1$--$64.1$ on Nova),
an order of magnitude below the spread between LLM backbones.
This is because embeddings only
surface candidate labels and a top-$K$ miss is recoverable. The pipeline is
also stable to input order: across three shuffled-order
seeds, NMI standard deviation is typically below $1$ point
and peaks at $\sim$$3$ on the hardest benchmarks
(Massive-D, Massive-I), well below the cross-LLM gap. Full
ablations in Appendix~\ref{app:more-ablation}.

\section{Production Validation}
\label{sec:production-validation}

RAILS is deployed as the clustering stage of a weekly
topic-discovery pipeline for customer-support tickets at
a SaaS support platform. We define a \emph{brand} as the
per-customer scope on the platform (e.g.\ a specific airline,
gaming studio, or food-delivery service), and process each
brand independently. A brand's first onboarding runs RAILS
against a historical backlog with
$\mathcal{L}^{(0)}{=}\emptyset$; subsequent weekly runs
process only newly-created tickets and seed the pool from
the brand's previous-run pool, so recurring topics keep
stable identities week-over-week while the LLM remains free
to introduce genuinely new themes.

\paragraph{Setup.}
We validate RAILS as a drop-in replacement for the prior
UMAP+HDBSCAN clustering stage along two complementary axes:
clustering \emph{quality} against an LLM-labelled reference set
(\S\ref{sec:prod-quality}), and \emph{operational behavior} at
production volume (\S\ref{sec:prod-operation}). Both axes use the
same four anonymised brands chosen to span unrelated industries
and traffic profiles (Brands~A--D). Both systems consume identical
pre-processed tickets and the same
\texttt{multilingual-e5-large-instruct}~\citep{wang2024multilingual}
encoder, fine-tuned in-house on customer-support intent pairs;
embeddings are pre-computed once upstream and reused by both
stacks, so any performance difference is attributable to
the clustering stage alone. The HDBSCAN baseline uses the fixed
hyperparameters that were tuned on a hold-out set.

We use Nova Lite as the LLM backbone: it trails Sonnet~4.5
by under $11$ NMI points on the intent-clustering tasks closest
in shape to support tickets (Massive-D, Massive-I, MTOP-I;
\S\ref{sec:results}) while clustering at $\sim$55$\times$
lower job cost for our token mix (Appendix~\ref{app:cost});
at weekly volume this trade-off is decisive.

\begin{table}[t]
\centering
\small
\setlength{\tabcolsep}{3pt}
\begin{tabular}{l rr rr rr rr}
\toprule
       & \multicolumn{2}{c}{Reference} & \multicolumn{2}{c}{ARI} & \multicolumn{2}{c}{NMI} & \multicolumn{2}{c}{Noise\,F1} \\
\cmidrule(lr){2-3}\cmidrule(lr){4-5}\cmidrule(lr){6-7}\cmidrule(lr){8-9}
Brand  & \#Lbl & N\% & R & H & R & H & R & H \\
\midrule
A food delivery   &   251 &  7.0 & \textbf{.44} & .29 & \textbf{.55} & .52 & \textbf{.59} & .17 \\
B social network  &   288 &  3.3 & \textbf{.14} & .06 & \textbf{.48} & .45 & \textbf{.45} & .06 \\
C gaming          &   316 & 11.1 & \textbf{.20} & .10 & \textbf{.49} & .48 & \textbf{.27} & .05 \\
D e-commerce      &   248 &  3.0 & \textbf{.17} & .09 & \textbf{.52} & .48 & \textbf{.50} & .03 \\
\midrule
Agg.              & 1{,}103 &  6.0 & \textbf{.28} & .05 & \textbf{.62} & .44 & \textbf{.43} & .06 \\
\bottomrule
\end{tabular}
\caption{Clustering quality of RAILS (\textbf{R}) vs.\
UMAP+HDBSCAN (\textbf{H}) on the internal reference set, 4k tickets per brand. \#Lbl is the
number of canonical reference labels and 
N\% is the share of reference \texttt{NOISE} tickets. RAILS values are means over three input-order seeds;
trial variance is small ($\sigma_{\mathrm{ARI}}{=}0.009$,
$\sigma_{\mathrm{NMI}}{=}0.002$).}
\label{tab:production-quality}
\end{table}

\begin{table}[t]
\centering
\small
\setlength{\tabcolsep}{3pt}
\begin{tabular}{l rr rr rr}
\toprule
       & \multicolumn{2}{c}{\#Topics} & \multicolumn{2}{c}{Noise\,\%} & \multicolumn{2}{c}{Wall\,(min)} \\
\cmidrule(lr){2-3}\cmidrule(lr){4-5}\cmidrule(lr){6-7}
Brand  & R & H & R & H & R & H \\
\midrule
A food delivery      &   534 & 1{,}297 & 6.7 &  8.2 & 19.0 & 5.4 \\
B social network     &   888 & 1{,}178 & 4.0 & 39.1 & 20.9 & 4.0 \\
C gaming     & 1{,}155 & 1{,}107 & 6.2 & 39.5 & 21.0 & 3.9 \\
D e-commerce     & 1{,}259 & 1{,}063 & 3.8 & 34.4 & 21.1 & 3.9 \\
\midrule
Mean         &   959 & 1{,}161 & 5.2 & 30.3 & 20.5 & 4.3 \\
Std.\ dev.   &   324 &   102 & 1.5 & 14.9 &  1.0 & 0.7 \\
\bottomrule
\end{tabular}
\caption{Operational behavior of RAILS (\textbf{R}) vs.\
UMAP+HDBSCAN (\textbf{H}) at production volume
($\approx$110k tickets per brand).
Topic counts: RAILS labels / HDBSCAN clusters. Noise\%:
\textsc{noise}-assigned (RAILS) or the $-1$ bucket (HDBSCAN).}
\label{tab:production}
\end{table}

\definecolor{brandA}{HTML}{2CA02C}
\definecolor{brandB}{HTML}{9467BD}
\definecolor{brandC}{HTML}{D62728}
\definecolor{brandD}{HTML}{FF7F0E}

\begin{figure}[t]
\centering
\begin{tikzpicture}
\pgfplotsset{
  every axis/.append style={
    width=0.95\columnwidth, height=5.6cm,
    font=\scriptsize,
    enlargelimits=false,
    xlabel={Tickets processed},
    xlabel style={font=\scriptsize, yshift=2pt},
    ylabel style={font=\scriptsize},
    tick label style={font=\tiny},
    x filter/.expression={x*15},
    scaled x ticks=base 10:-3,
    xtick scale label code/.code={},
    xticklabel={\pgfmathprintnumber{\tick}k},
    xmin=0,
  },
}

\begin{axis}[
  name=main,
  axis y line*=left,
  axis x line*=bottom,
  ylabel={Label pool size $\lvert\mathcal{L}\rvert$},
  ylabel style={yshift=-3pt},
  ymin=0, ymax=1500,
  ytick={0,500,1000,1500},
  ymajorgrids, grid style={dotted, gray!40},
  legend style={
    at={(0.02,0.97)}, anchor=north west,
    font=\tiny,
    draw=gray!40, line width=0.3pt,
    fill=white, fill opacity=0.92, text opacity=1,
    row sep=0pt, inner xsep=4pt, inner ysep=3pt,
    nodes={inner sep=1pt},
  },
  legend image post style={line width=1.2pt},
  legend cell align=left,
]
\addplot+[mark=none, line width=1.0pt, color=brandA] table[x=batch, y=pool_size] {figures/data/pool_brandA.dat};
\addlegendentry{A\, food delivery}
\addplot+[mark=none, line width=1.0pt, color=brandB] table[x=batch, y=pool_size] {figures/data/pool_brandB.dat};
\addlegendentry{B\, social network}
\addplot+[mark=none, line width=1.0pt, color=brandC] table[x=batch, y=pool_size] {figures/data/pool_brandC.dat};
\addlegendentry{C\, gaming}
\addplot+[mark=none, line width=1.0pt, color=brandD] table[x=batch, y=pool_size] {figures/data/pool_brandD.dat};
\addlegendentry{D\, e-commerce}
\end{axis}

\begin{axis}[
  axis y line*=right,
  axis x line=none,
  ylabel={New labels / batch},
  ylabel style={yshift=6pt},
  ymin=0, ymax=15,
  ytick={0,5,10,15},
  legend style={
    at={(0.98,0.97)}, anchor=north east,
    font=\tiny,
    draw=gray!40, line width=0.3pt,
    fill=white, fill opacity=0.92, text opacity=1,
    row sep=0pt, inner xsep=4pt, inner ysep=3pt,
    nodes={inner sep=1pt},
  },
  legend image post style={line width=1.0pt},
  legend cell align=left,
]
\addlegendimage{black!75, line width=1.0pt}
\addlegendentry{solid: pool size $\lvert\mathcal{L}\rvert$}
\addlegendimage{black!75, dashed, line width=0.7pt}
\addlegendentry{dashed: new / batch}

\addplot[dashed, line width=0.55pt, color=brandA, mark=none, opacity=0.75] table[x=batch, y=new_labels] {figures/data/pool_brandA.dat};
\addplot[dashed, line width=0.55pt, color=brandB, mark=none, opacity=0.75] table[x=batch, y=new_labels] {figures/data/pool_brandB.dat};
\addplot[dashed, line width=0.55pt, color=brandC, mark=none, opacity=0.75] table[x=batch, y=new_labels] {figures/data/pool_brandC.dat};
\addplot[dashed, line width=0.55pt, color=brandD, mark=none, opacity=0.75] table[x=batch, y=new_labels] {figures/data/pool_brandD.dat};
\end{axis}

\end{tikzpicture}

\caption{Label pool convergence across four production brands; the first $100$k tickets per brand are shown for visual comparability (full runs in Table~\ref{tab:production}). Solid lines (left axis) show cumulative pool size $\lvert\mathcal{L}\rvert$; dashed lines (right axis) show new labels appended per batch. Per-batch generation decays within the first few thousand tickets as retrieval surfaces existing labels.}
\label{fig:pool-growth}
\end{figure}
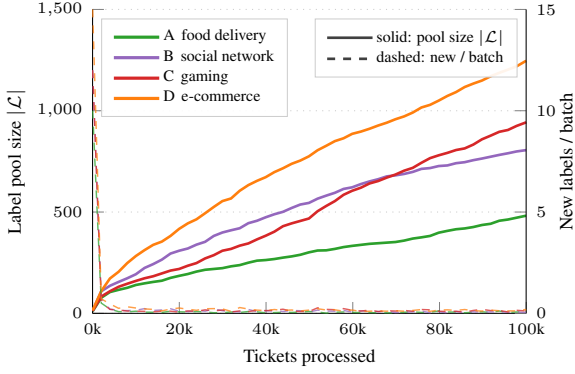

\subsection{Clustering Quality}
\label{sec:prod-quality}

Table~\ref{tab:production-quality} reports clustering quality against
an internal reference set that is constructed by an independent LLM (Claude 4.5 Sonnet) with a deliberately different prompt and a different architecture (separate labeling and merging stages, no retrieval, no pool). We detail these independence axes in Appendix~\ref{app:inhouse_benchmark_creation} to limit LLM-as-judge bias.

RAILS improves on
every brand and every metric. Per-brand ARI rises by $8$--$15$
percentage points and NMI by $1$--$4$ points; trial variance over
three input-order seeds is small
($\sigma_{\mathrm{ARI}}{=}0.009$,
$\sigma_{\mathrm{NMI}}{=}0.002$), indicating the streaming pipeline
is order-stable.

RAILS also routes noise far more accurately: aggregate
Noise\,F1 of $0.43$ vs.\ $0.06$ for HDBSCAN's $-1$ bucket, a
$7\times$ gap that holds on every brand ($3$--$18\times$
per-brand). This is consistent with the LLM applying the
prompt's explicit content-free definition uniformly, whereas
HDBSCAN's $-1$ bucket piles up low-density tickets,
sparse-region embeddings, and genuinely content-free inputs.

\subsection{Operational Behavior}
\label{sec:prod-operation}

Table~\ref{tab:production} reports operational behavior on the same
four brands at production volume ($\approx$110k tickets each,
$438{,}759$ total), simulating a high-volume ingestion
window.

\paragraph{Coverage and topic counts.}
RAILS retains $93$--$96\%$ of tickets in real topics across all
four brands, whereas HDBSCAN drops $34$--$40\%$ of tickets into
the $-1$ bucket for three of the four brands (B, C, D). Under
fixed prompt-level granularity RAILS produces $534$--$1{,}259$
labels per brand; HDBSCAN produces $1{,}063$--$1{,}297$ clusters
under fixed hyperparameters. RAILS' counts track interpretable
brand breadth: e-commerce ($1{,}259$) and gaming ($1{,}155$) serve
broad product catalogs and produce roughly twice as many topics
as food delivery ($534$), whose request mix is narrower, with
social network's ($888$) sitting between. HDBSCAN's output is nearly
flat across brands.

\paragraph{Throughput.}
RAILS is $4$--$5\times$ slower than HDBSCAN in absolute wall
time ($20.5$ vs.\ $4.3$~min per brand)
because the LLM call is now on the critical path; in production
terms this is comfortably within our weekly SLA. Wall time scales
near-linearly with ticket count at $\sim$$89$~tickets/s end-to-end,
with per-batch LLM latency averaging $1.48$--$1.87$~s. Across
$\sim$$29{,}250$ LLM calls we observed zero failures and zero
retries; the slow-start controller (\S\ref{sec:concurrency}) ramped
from $1$ to $W_{\max}{=}15$ linearly for every brand, with no
provider throttling.

\subsection{Label Pool Dynamics}
\label{sec:prod-pool}

Figure~\ref{fig:pool-growth} shows the label pool growth rate for
different brands. The LLM directly reuses a retrieved label for
$\sim$$90\%$ of tickets (range $87.9$--$91.5\%$, $\sigma{=}1.5$),
and $\sim$$78\%$ of newly-coined labels are absorbed by the merge
step in~\S\ref{sec:merge} (range $67.3$--$82.9\%$). Per-batch
label generation decays within the first few thousand tickets as
retrieval saturates, and pool sizes plateau between $\sim$$0.5$k
and $1.3$k labels per brand. The $\sim$$90\%$ in-run reuse rate is also a direct representation of
steady-state behavior under weekly operation.

\section{Conclusion}
We presented \textsc{RAILS}, a retrieval-augmented incremental
LLM clustering method designed for production deployment.
By combining top-$K$ retrieval from a persistent label pool,
batched LLM assignment with slow-start adaptive concurrency,
and embedding-space label merging, \textsc{RAILS} matches or
exceeds prior LLM-clustering work on six public benchmarks. We
further demonstrated production-scale behavior on a
representative slice of customer support tickets spanning four
industries, yielding substantially cleaner topic structure than
the HDBSCAN pipeline it replaced. We expect the design to
generalize to any setting where (i)~robust LLM-based text
understanding is preferred,
(ii)~per-document throughput matters, and (iii)~operators need
a natural-language control surface over cluster semantics. We
hope the design choices reported here offer a practical recipe
for teams deploying LLM clustering at scale.

\section*{Ethical Considerations}

Support tickets contain user-generated text that may include personal information. All processing in this work was performed on internal infrastructure with the data-handling controls required by the deploying organization. The LLM provider operated under a zero-retention and no-training-on-data agreement. We remove Personally Identifiable Information (PII) from tickets (including name, e-mail and mailing address) before sending the data to the LLM.

\section*{Acknowledgments}
We thank André Nascimento, Sebastian Niemczyk, Ionut Militaru and the rest of the team for discussions and support in bringing the system described in this work into production.

\section*{Limitations}
\label{limitations}

The slow-start controller adapts well-behaved concurrency to provider
capacity, but the ceiling on throughput is set by provider rate limits on tokens and requests per minute. In practice this means
cold-start runs on large new brands can still take hours, and a
provider outage stalls the entire pipeline. Self-hosted deployments
shift the ceiling from quotas to GPU budget but do not remove it.

The architecture also treats prompts as the primary control surface
for semantics, granularity, and noise handling, which places a hard
dependency on the LLM's instruction-following capability. Smaller or
weaker models may violate the prompt contract; e.g. we have
observed non-English tickets misclassified as \textsc{noise}.

The label pool is append-only by design: once a label is
admitted it persists across all subsequent runs, and we never split,
retire, or rename labels in place. This maximises topic stability for
downstream products that expose topics to end users. Use cases that
require tracking concept drift, reorganising stale labels, or merging
and splitting topics as a brand's product evolves would need to be
layered on top; for example, as a periodic maintenance pass over the
pool.

The slow-start concurrency controller (\S\ref{sec:concurrency})
is supported by observational evidence rather than a controlled
ablation. 

Production tickets contain customer PII and are not re-distributable; full source code is similarly restricted. We compensate with public-benchmark results that any reader can reproduce, with complete prompts and hyperparameters.

\bibliography{custom}

\appendix

\section{In-house benchmark}
\label{app:inhouse_benchmark_creation}

We construct an LLM-annotated reference set following the approach of
\citet{huang2025}, adapted to avoid the expensive classification pass.
Instead of a generate--merge--classify pipeline, we track ticket-to-label
assignments during generation and label-to-label mappings during merging,
then remap assignments transitively.

\paragraph{Tickets.}
We sample up to $4{,}000$ \emph{email} tickets per brand from the
same four anonymised brands used in the production evaluation
(\S\ref{sec:production-validation}); per-brand reference
label statistics are in Table~\ref{tab:production-quality}. Tickets are preprocessed
identically to the production pipeline: text cleaning followed by
end-user-only message concatenation.

\paragraph{Label generation.}
We use the \emph{open-ended} reference: \textit{Claude~4.5~Sonnet} processes
tickets in mini-batches of $B{=}50$ under a minimum-instruction prompt
(verbatim):
\begin{quote}\small\itshape
Below are support tickets. Group them by topic and give each group a short
descriptive name. Decide the right level of granularity yourself based on
what you see. If a ticket has no real content to classify, label it
\texttt{NOISE}.
\end{quote}
For each batch the model proposes candidate topic labels and simultaneously
assigns each ticket to one label.

\paragraph{Iterative label merging.}
\textit{Claude~4.5~Sonnet} consolidates semantically similar labels and
returns an old$\to$new mapping. When the label set exceeds $K{=}200$, it is
split into chunks of at most $K$ labels and each chunk is merged
separately. Merging runs as an iterative loop (up to $I_{\max}{=}3$
iterations) that terminates on the first satisfied condition:
(i)~the label set fits in a single call ($N \leq K$),
(ii)~reduction falls below 5\% of the prior set size,
(iii)~the canonical set repeats a previously seen state (oscillation
guard), or (iv)~the iteration cap is reached.

\paragraph{Independence from RAILS.}
\label{par:refset_independence}
This reference is constructed independently of RAILS along three axes,
limiting the risk of LLM-as-judge bias inflating RAILS's reported scores:
\begin{description}
  \item[Prompt.] RAILS uses a granularity-anchored triage prompt with
    cross-domain examples (\emph{``Refund for Damaged Item''} rather than
    \emph{``Order Issues''}; see Appendix~\ref{app:internal_prompts}). The
    reference uses the minimum-instruction prompt shown above: no examples,
    no triage framing, no granularity prior. Neither prompt informs the
    other.
  \item[Clustering architecture.] RAILS is single-pass incremental: each
    mini-batch sees a top-$K$ retrieval over a growing per-brand pool,
    followed by an embedding-space merge. The reference performs monolithic
    in-batch labelling (each call sees $B{=}50$ tickets independently, no
    retrieval, no pool) followed by iterative cross-batch label merging --
    a fundamentally different procedure that exercises different failure
    modes.
  \item[Model class.] The reference uses Claude~4.5~Sonnet. The deployed RAILS
    evaluation uses Nova~Lite~v1 -- a weaker model from a different family
    that scores~11 NMI points below Sonnet on the public benchmarks
    (Table~\ref{tab:public_results}).

\end{description}

\section{Deployment Configuration}
\label{app:deployment}

\begin{table}[t]
\centering
\footnotesize
\setlength{\tabcolsep}{4pt}
\renewcommand{\arraystretch}{1.0}
\resizebox{\columnwidth}{!}{%
\begin{tabular}{@{}lr@{}}
\toprule
\textbf{Parameter} & \textbf{Value} \\
\midrule
\multicolumn{2}{@{}l}{\emph{Shared encoder}} \\
Base model                            & \texttt{ml-e5-large-instruct} \\
Fine-tuning                           & in-house contrastive \\
Embedding dim.\ / pooling             & 1024 / mean \\
Max tokens                            & 512 \\
\midrule
\multicolumn{2}{@{}l}{\emph{Baseline: UMAP + HDBSCAN}} \\
UMAP \texttt{n\_components}, \texttt{n\_neighbors} & 16, 15 \\
UMAP metric / init                    & cosine / spectral \\
HDBSCAN impl.\ / metric / algorithm   & \texttt{sklearn} / euclidean / \texttt{brute} \\
HDBSCAN \texttt{min\_cluster\_size}   & 5 \\
HDBSCAN \texttt{selection\_epsilon}   & 0.1 \\
\midrule
\multicolumn{2}{@{}l}{\emph{RAILS}} \\
LLM                                   & Nova Lite v1 \\
Temperature / output cap              & 0 / 700 tokens \\
Batch size $B$ / top-$K$              & 15 / 5 \\
Merge threshold $\tau$                & 0.85 \\
Concurrency $W_{\min},\,W_{\max}$     & 1,\,15 \\
\bottomrule
\end{tabular}}
\caption{Production configuration of both clustering stacks.}
\label{tab:deployment-config}
\end{table}

This appendix details the exact production configuration of both
clustering stacks compared in \S\ref{sec:production-validation}.
Hyperparameters were tuned on an internal validation set and held
fixed across all brands and experiments. All experiments run on AWS g5.16xlarge instances (64 vCPUs, 256\,GB RAM), except the Gemma~4 26B MoE public benchmark runs, which used a g5.12xlarge instance (4$\times$ A10G GPUs, 48\,vCPUs, 192\,GB RAM) to serve the model via vLLM.

Table~\ref{tab:deployment-config} lists all hyperparameters; the embedder, \texttt{multilingual-e5-large-instruct}%
~\citep{wang2024multilingual}, is fine-tuned in-house on
ticket-intent pairs with a soft-contrastive objective. Its 1024-d
output is shared across stacks so any
quality difference is attributable to the clustering stage alone.
Embeddings are computed once by an upstream weekly job and cached,
so neither stack pays encoding cost at clustering time.
UMAP~\citep{mcinnes2018umap} reduces the embeddings to 16
dimensions before HDBSCAN~\citep{campello2013density}
(\texttt{sklearn} implementation) extracts variable-density
clusters.

\section{Cost}
\label{app:cost}

We estimate end-to-end LLM cost using the production token shape
observed during the runs reported in §\ref{sec:production-validation}. Each
batched call to the LLM consumes
on average $\sim$1{,}705 input tokens (400 system + 80 user-template
+ $B{=}15$ tickets at $\sim$55 tokens each + up to $75$ retrieved
labels at $\sim$10 tokens each) and produces only $\sim$250 output
tokens (assignment dictionary). At $B{=}15$,
a 100k-ticket brand therefore issues $\approx$6{,}667 generation calls,
amounting to $\sim$11.4\,M input and $\sim$1.67\,M output tokens.

Because hosted LLM pricing is dominated by output tokens
(typically $4$--$5\times$ the input rate), our design deliberately
shifts work onto the input side: the retrieved candidate set and
batched ticket payload inflate input tokens, but the structured
JSON response stays compact with one short label per ticket. The $\approx$7:1 input-to-output
ratio is what makes per-ticket cost competitive even with a
frontier-class model.

\begin{table}[h]
\centering
\small
\setlength{\tabcolsep}{4pt}
\begin{tabular}{lrrr}
\toprule
LLM & In (\$/MTok) & Out (\$/MTok) & Cost / 100k \\
\midrule
Nova Lite      & 0.06 & 0.24  & \$1.08  \\
Haiku 4.5      & 1.00 & 5.00  & \$19.72 \\
Sonnet 4.5     & 3.00 & 15.00 & \$59.16 \\
\bottomrule
\end{tabular}
\caption{Estimated LLM cost for clustering 100k tickets at $B{=}15$,
$K{=}5$, assuming the token shape described above.
Pricing reflects AWS Bedrock on-demand list prices at submission time;
batch-inference or provisioned-throughput pricing would lower these
figures further. Embedding and other operation costs are
excluded.}
\label{tab:cost}
\end{table}

Gemma~4 26B MoE is excluded from Table \ref{tab:cost}; as a self-hosted model (served via vLLM), its cost depends on the GPU instance provisioned rather than a per-token API rate.

\section{Benchmark datasets}
\label{app:datasets}

\begin{table}[h]
\centering
\small
\begin{tabular}{lrrl}
\toprule
Dataset & Samples & Clusters & Task \\
\midrule
\texttt{arxiv}     & 3{,}674 &  93 & Paper subject areas \\
\texttt{MTOP-I}    & 4{,}386 & 102 & Conversational intents \\
\texttt{Massive-I} & 2{,}974 &  59 & Conversational intents \\
\texttt{Massive-D} & 2{,}974 &  18 & Scenario-level domains \\
\texttt{few\_nerd} & 3{,}789 &  58 & Entity types \\
\texttt{few\_rel}  & 4{,}480 &  64 & Semantic relations \\
\bottomrule
\end{tabular}
\caption{Public benchmarks used in §\ref{sec:results}.
Splits from~\citet{clusterllm}.}
\label{tab:datasets}
\end{table}

\section{Ablation: embedding model, LLM, and input order}
\label{app:more-ablation}

We complement the headline numbers in Table~\ref{tab:public_results}
with a full grid over multiple embedding models and LLMs from
different families and price points. All cells in
Table~\ref{tab:ablation} use the same prompts, batch size, and merge
threshold as the main experiments (\S\ref{sec:setup}); only the
embedding model and LLM swap. Standard deviations are calculated over 3 independent runs with different random shufflings to measure robustness to input ordering.

\begin{table*}[t]
\centering
\resizebox{\textwidth}{!}{
\begin{tabular}{ll|ccc|ccc|ccc|ccc}
\toprule
& & \multicolumn{3}{c|}{\textbf{Sonnet 4.5}} & \multicolumn{3}{c|}{\textbf{Haiku 4.5}} & \multicolumn{3}{c}{\textbf{Nova Lite}} & \multicolumn{3}{c}{\textbf{Gemma 4 26B MoE}} \\
\cmidrule(lr){3-5} \cmidrule(lr){6-8} \cmidrule(lr){9-11} \cmidrule(lr){12-14}
\textbf{Dataset} & \textbf{Embedding} & ACC & NMI & ARI & ACC & NMI & ARI & ACC & NMI & ARI & ACC & NMI & ARI\\
\midrule
\multirow{3}{*}{ArxivS2S}
& MPNet & 32.5±0.6 & 62.9±0.5 & 20.1±0.4 & 28.3±0.5 & 56.2±0.2 & 15.1±1.1 & 23.8±1.9 & 50.4±0.9 & 11.8±1.2 & 33.1±0.6 & 58.4±0.6 & 19.7±0.4 \\
& Instructor & 30.8±0.3 & 60.4±0.4 & 17.7±0.4 & 27.7±1.3 & 54.7±1.0 & 14.3±1.5 & 23.3±1.4 & 49.2±0.6 & 11.0±0.9 & 30.0±0.7 & 56.1±0.4 & 17.2±0.4\\
& Intent & 31.2±0.8 & 62.5±1.0 & 19.3±0.7 & 26.8±1.9 & 56.6±1.0 & 14.4±0.2 & 22.9±1.6 & 50.1±0.6 & 10.7±1.3 & 31.6±1.2 & 57.6±0.4 & 18.7±0.4\\
\midrule
\multirow{3}{*}{MTOP-I}
& MPNet & 76.2±0.3 & 85.1±0.3 & 84.1±0.2 & 71.2±1.7 & 81.3±0.2 & 78.5±0.6 & 64.9±3.8 & 76.4±1.1 & 66.2±7.1 & 71.7±1.1 & 80.2±0.4 & 76.0±1.7\\
& Instructor & 76.2±0.6 & 85.0±0.2 & 82.4±1.3 & 71.9±0.9 & 81.5±1.0 & 78.9±1.1 & 64.8±3.8 & 76.2±1.2 & 66.5±6.7 & 71.7±1.5 & 80.2±1.2 & 73.0±4.9\\
& Intent & 76.3±1.3 & 85.0±0.5 & 83.6±1.1 & 72.5±0.6 & 80.7±0.7 & 78.9±0.5 & 65.8±2.7 & 76.1±1.5 & 66.2±5.4 & 70.3±3.3 & 78.6±1.9 & 65.9±10.1\\
\midrule
\multirow{3}{*}{Massive-I}
& MPNet & 60.4±1.3 & 77.5±0.7 & 54.3±1.1 & 56.2±0.8 & 72.6±1.6 & 45.9±1.3 & 47.8±4.2 & 66.8±1.5 & 38.2±1.8 & 54.4±1.8 & 70.8±1.4 & 44.8±1.1\\
& Instructor & 61.2±2.8 & 77.3±2.3 & 54.7±2.4 & 56.4±1.6 & 72.6±1.7 & 45.0±2.2 & 45.6±3.3 & 64.5±2.8 & 34.9±4.3 & 54.8±0.7 & 70.0±1.3 & 40.2±1.1\\
& Intent & 60.0±3.4 & 77.3±1.5 & 54.8±1.4 & 57.8±1.1 & 73.7±1.2 & 46.8±1.5 & 47.3±3.5 & 66.4±0.3 & 36.6±3.8 & 55.5±0.9 & 71.2±0.7 & 43.3±2.7\\
\midrule
\multirow{3}{*}{Massive-D}
& MPNet & 70.6±2.0 & 76.4±2.9 & 60.8±3.6 & 66.4±0.5 & 70.3±1.8 & 56.6±1.4 & 65.5±0.9 & 67.6±1.8 & 56.5±1.9 & 62.5±1.6 & 68.3±1.1 & 55.2±1.6\\
& Instructor & 71.9±2.8 & 77.0±3.0 & 61.7±3.7 & 62.6±1.4 & 68.1±2.7 & 52.7±2.6 & 65.6±1.3 & 65.8±1.8 & 56.1±2.1 & 62.7±0.8 & 67.6±0.7 & 53.6±0.9\\
& Intent & 72.2±2.4 & 77.0±2.7 & 62.0±3.1 & 66.6±0.4 & 70.4±1.2 & 56.0±0.6 & 65.7±1.8 & 67.0±1.8 & 56.4±2.7 & 64.2±0.9 & 69.6±0.8 & 56.2±0.5\\
\midrule
\multirow{3}{*}{FewNerd}
& MPNet & 60.0±1.7 & 72.7±0.4 & 62.5±2.4 & 63.8±1.2 & 70.9±0.6 & 66.3±5.6 & 58.4±1.4 & 64.1±1.8 & 55.9±1.0 & 65.1±2.2 & 71.5±0.3 & 66.3±5.3\\
& Instructor & 64.3±1.4 & 73.8±0.5 & 64.6±3.9 & 63.4±1.6 & 70.6±0.3 & 65.3±7.3 & 58.6±0.8 & 64.2±1.3 & 55.9±4.5 & 65.2±1.4 & 71.3±0.3 & 67.6±2.5\\
& Intent & 61.1±1.1 & 73.4±0.5 & 63.0±2.7 & 63.4±0.5 & 70.1±0.6 & 64.4±6.2 & 57.1±2.9 & 63.5±1.4 & 52.4±4.1 & 65.9±0.3 & 71.6±0.6 & 69.4±3.2\\
\midrule
\multirow{3}{*}{FewRel}
& MPNet & 56.1±0.2 & 74.3±0.1 & 46.2±0.8 & 38.6±1.4 & 64.1±0.8 & 26.7±1.7 & 32.1±0.6 & 59.2±0.5 & 20.3±1.0 & 41.5±1.7 & 65.2±0.6 & 27.0±1.8\\
& Instructor & 51.9±2.0 & 72.8±0.6 & 40.8±1.6 & 38.0±1.5 & 63.8±0.5 & 25.4±1.7 & 32.5±1.3 & 58.5±1.2 & 19.5±1.0 & 40.2±1.7 & 64.8±0.7 & 24.6±1.3\\
& Intent & 55.1±1.1 & 73.8±1.0 & 44.5±0.9 & 39.9±1.1 & 64.2±1.2 & 26.6±2.2 & 32.3±1.1 & 58.4±0.8 & 19.1±0.7 & 42.4±1.8 & 65.7±0.8 & 25.8±1.1\\
\midrule
\multirow{3}{*}{\textit{Average}}
& MPNet & 59.3 & 74.8 & 54.7 & 54.1 & 69.2 & 48.2 & 48.8 & 64.1 & 41.5 & 54.7 & 69.1 & 48.2\\
& Instructor & 59.4 & 74.4 & 53.7 & 53.3 & 68.5 & 46.9 & 48.4 & 63.1 & 40.6 & 54.1 & 68.3 & 46.0\\
& Intent & 59.3 & 74.8 & 54.5 & 54.5 & 69.3 & 47.9 & 48.5 & 63.6 & 40.2 & 55.0 & 69.1 & 46.6\\
\bottomrule
\end{tabular}
}
\caption{Ablation across embedding models and LLMs on six
public benchmarks. Mean $\pm$~std over three independent runs
with shuffled input order; std reflects both LLM decoding
non-determinism and batch-composition variance. LLM choice
drives variation; embedding model and input order have near-zero
effect (\S\ref{app:more-ablation}).}
\label{tab:ablation}
\end{table*}

\paragraph{Findings.}
Three patterns are robust across all six benchmarks:

\begin{itemize}
\item \textbf{LLM choice dominates.} Within each embedding column,
swapping the LLM moves average NMI by several points and average
ARI by roughly an order of magnitude more than swapping the
embedder. The ordering of LLMs is preserved on every embedding,
with the largest gaps appearing on fine-grained semantic tasks
(FewRel, MTOP-I) where stronger reasoning helps disambiguate
subtle distinctions.

\item \textbf{Embedding choice is nearly irrelevant.} Holding the LLM fixed, 
average NMI varies by at most $\sim$1.0 point across embedders. 
This is by design (\S\ref{sec:retrieval}): embeddings
only surface candidate labels for the LLM, and a top-$K$
retrieval miss is recoverable because the LLM remains the final
arbiter, free to invent a new label when none of the retrieved
ones fits. The result makes the method practical to deploy without
an embedding-model tuning loop.

\item \textbf{Input order is irrelevant in practice.} The three
runs per cell shuffle ticket order across batches with independent
random seeds, so each cell measures both LLM-decoding non-determinism
and batch-composition variance. Standard deviations remain small on 
the easier benchmarks ($\sigma_{\mathrm{NMI}}{\leq}1.0$ on ArxivS2S, 
${\leq}1.9$ on MTOP-I and FewNerd) and grow only modestly on the 
harder ones (Massive-I, Massive-D peak at $\sigma_{\mathrm{NMI}}{\approx}3.0$, 
still well below cross-LLM gaps). The streaming, online label-pool
construction is therefore stable under reordering despite never
seeing the full dataset at once.
\end{itemize}

\section{Prompts}

\subsection{Internal Customer Support}
\label{app:internal_prompts}

\paragraph{System prompt}
\begin{quote}\small
You are an expert at categorizing customer support tickets into meaningful topic labels.

Given a batch of support tickets, do TWO things: (1) identify the distinct topic categories these tickets belong to; (2) assign each ticket to exactly one of those categories.

Each label should be a short, descriptive phrase (3--8 words) that captures the core issue or request type.

\textbf{Granularity.} Aim for the granularity of a support team's triage categories: specific enough that an agent immediately knows the domain and action required, but general enough to group tickets with the same root cause or request type. Examples across industries: ``Refund for Damaged Item'' (not ``Order Issues''), ``SSO Login Error'' (not ``Product Bug''), ``Flight Cancellation Refund'' (not ``Booking Issue''), ``Prescription Refill Request'' (not ``Appointment Related'').

\textbf{Rules.} Every ticket MUST be assigned to exactly one label. Multiple tickets can share a label. Assign \texttt{NOISE} \emph{only} for tickets that are truly unclassifiable: automated system messages, completely empty content, or pure gibberish with zero discernible intent. Do NOT label as NOISE short messages, non-English text, vague requests, anonymized placeholders, or tickets where you can infer any plausible support intent. When in doubt, assign a topic label. Return ONLY a JSON object with the requested keys.
\end{quote}

\paragraph{User prompt template}
\begin{quote}\small
Here are \{n\_tickets\} support tickets. Identify topic categories and assign each ticket.

\{existing\_block\}

\{tickets\_text\}

Return JSON with EXACTLY these two keys, in this order: \texttt{assignments} (label per ticket; prefer an existing category) and \texttt{new\_labels} (DISTINCT list of newly introduced labels referenced by \texttt{assignments}, $\leq M_{\text{new}}$).
\end{quote}

\subsection{Public Benchmarks}
\label{app:public_prompts}

For each public benchmark dataset, we adapted the prompt to match the domain and desired label granularity. Below are the system and user prompts for the best-performing configuration on each dataset. All prompts share a common output format specification (shown below) appended to the user prompt.

\paragraph{Common output format (appended to all user prompts)}
\begin{quote}\small
Return JSON with EXACTLY these two keys, in this order:

\begin{enumerate}
\item \texttt{assignments}---for each of the \{n\_tickets\} tickets, the topic label you chose (prefer an existing category; only create a new one when no existing category fits).
\item \texttt{new\_labels}---the DISTINCT list of any newly introduced labels referenced by \texttt{assignments}. Do NOT repeat labels. Do NOT include existing categories here. If every ticket fits an existing category, return [].
\end{enumerate}

\textbf{HARD LIMITS (critical):}
\begin{itemize}
\item \texttt{new\_labels} must contain AT MOST 5 entries, all distinct.
\item Every string in \texttt{new\_labels} must appear in \texttt{assignments}.
\item Do NOT repeat or paraphrase labels across entries.
\end{itemize}
{\scriptsize
\begin{verbatim}
{
  "assignments": {
    "Ticket 1": "label_used_for_ticket_1",
    "Ticket 2": "label_used_for_ticket_2",
    ...
    "Ticket {N}": "label_used_for_ticket_{N}"
  },
  "new_labels": ["new_label_1", ...]
}
\end{verbatim}
}
\end{quote}

\subsubsection{ArxivS2S}

\paragraph{System prompt}
\begin{quote}\small
You are an expert at organizing scientific papers into broad academic subject categories. Your task is to identify the subject area like e.g. ``Information Theory'', ``Combinatorics'', ``Group Theory'', ``Plasma Physics'' that each paper belongs to based on its title.
\end{quote}

\paragraph{User prompt}
\begin{quote}\small
Here are \{n\_tickets\} scientific paper titles. Identify the academic subject category for each paper.

\textbf{Guidelines:}
\begin{itemize}
\item Use subject categories like ``Graph Theory'', ``Quantum Physics'', etc.
\item Papers in the same field belong together even if studying different specific problems
\item Do NOT fragment by specific research problem---papers on ``Cell Multiplication'' and ``Cell Differentiation'' should both be ``Cell Biology''
\item Use standard academic subject taxonomy
\item If a paper title is too vague to categorize, assign ``NOISE''
\end{itemize}

\{existing\_block\}

\{tickets\_text\}

[Common output format appended here]
\end{quote}

\subsubsection{MTOP-I}

\paragraph{System prompt}
\begin{quote}\small
You are an expert at classifying user intents in task-oriented dialogue systems. Your task is to identify the intent for each user utterance, respecting meaningful distinctions between different user goals. When in doubt between similar intents, use the broader/more established category.
\end{quote}

\paragraph{User prompt}
\begin{quote}\small
Here are \{n\_tickets\} user utterances. Identify the intent category for each utterance.

\textbf{Guidelines for grouping:}
\begin{itemize}
\item Utterances with the same goal or action belong together even if phrased differently or if specific actions differ slightly
\item Use clear intent names (e.g., ``GET\_WEATHER'', ``SET\_ALARM'', ``PLAY\_MUSIC'')
\item Avoid fragmenting similar intents (e.g., ``weather\_today'' and ``weather\_forecast'', both are ``GET\_WEATHER'')
\item Intents should be actionable and distinct
\item If an utterance is unclear or ambiguous, assign ``NOISE''
\end{itemize}

\{existing\_block\}

\{tickets\_text\}

[Common output format appended here]
\end{quote}

\subsubsection{Massive-I}

\paragraph{System prompt}
\begin{quote}\small
You are an expert at classifying user intents in a massive, diverse conversational AI system. Your task is to identify and assign each user utterance to the appropriate intent category.
\end{quote}

\paragraph{User prompt}
\begin{quote}\small
Here are \{n\_tickets\} user utterances. Identify the type of intent each utterance expresses.

\textbf{Guidelines for grouping:}
\begin{itemize}
\item Utterances with the same goal or action belong together even if phrased differently
\item Within structured domains (alarms, datetime, playing music, etc.), preserve meaningful action distinctions: QUERY $\neq$ REMOVE $\neq$ SET $\neq$ CONVERT
\item Use e.g. ``general greet'' or ``general joke'' as valid catch-alls for non-actionable, or very vague utterances within a domain, rather than creating new intent categories
\item Avoid fragmenting similar intents (e.g., ``weather\_today'' and ``weather\_forecast'', both are ``GET\_WEATHER'')
\item Intents should be actionable and distinct
\item If an utterance is unclear or ambiguous, assign ``NOISE''
\end{itemize}

\{existing\_block\}

\{tickets\_text\}

[Common output format appended here]
\end{quote}

\subsubsection{Massive-D}

\paragraph{System prompt}
\begin{quote}\small
You are an expert at classifying user requests into broad application scenarios. Your task is to assign each user utterance to the scenario/domain it belongs to.

\textbf{Few-Shot Examples:}
\begin{enumerate}
\item Input: ``forward this message to my team at work'' $\to$ Category: ``email''
\item Input: ``post a picture of my lunch on instagram'' $\to$ Category: ``social''
\item Input: ``who directed the movie avatar'' $\to$ Category: ``qa''
\item Input: ``compose a new message and send it to sarah'' $\to$ Category: ``email''
\item Input: ``show me restaurants nearby that have outdoor seating'' $\to$ Category: ``recommendation''
\item Input: ``what's the capital of france'' $\to$ Category: ``qa''
\item Input: ``find me fun activities to do this weekend'' $\to$ Category: ``recommendation''
\item Input: ``move these emails to the archive folder'' $\to$ Category: ``email''
\item Input: ``that's silly'' $\to$ Category: ``general''
\end{enumerate}
\end{quote}

\paragraph{User prompt}
\begin{quote}\small
Here are \{n\_tickets\} user utterances. Identify the conversation scenario or domain for each utterance.

\textbf{Guidelines for grouping:}
\begin{itemize}
\item Focus on the general scenario domain (e.g., ``alarm'', ``weather'', ``calendar'', ``music'')
\item Utterances in the same scenario domain belong together regardless of specific intent
\item Don't fragment by action within a domain
\item Use simple, broad scenario names
\item Group by domain, not by specific action
\item If an utterance is trying to access one domain but is poorly phrased, use that domain
\item If a scenario is truly unclear, assign ``NOISE''
\end{itemize}

\{existing\_block\}

\{tickets\_text\}

[Common output format appended here]
\end{quote}

\subsubsection{FewNerd}

\paragraph{System prompt}
\begin{quote}\small
You are an expert at Named Entity Recognition using appropriate category breadth. Your task is to classify entity types based on context.

\textbf{Few-Shot Examples:}
\begin{enumerate}
\item Text: ``Paris hosted the Olympics in 1900. The entity type of Paris.'' $\to$ Type: ``location''
\item Text: ``Michael Jordan led the Bulls to six championships. The entity type of Michael Jordan.'' $\to$ Type: ``athlete''
\item Text: ``Google announced new features yesterday. The entity type of Google.'' $\to$ Type: ``company''
\item Text: ``The Nile River flows through eleven countries. The entity type of Nile River.'' $\to$ Type: ``water body''
\item Text: ``Senator Warren proposed new legislation. The entity type of Senator Warren.'' $\to$ Type: ``politician''
\item Text: ``Harvard University has 22,000 students. The entity type of Harvard University.'' $\to$ Type: ``education''
\item Text: ``Chinese traditions date back millennia. The entity type of Chinese.'' $\to$ Type: ``location''
\item Text: ``The hemoglobin protein carries oxygen. The entity type of hemoglobin.'' $\to$ Type: ``biological entity''
\end{enumerate}
\end{quote}

\paragraph{User prompt}
\begin{quote}\small
Here are \{n\_tickets\} texts. Each ends with ``The entity type of X.'' Identify the type for entity X based on the context, using broad categories when appropriate but distinguishing specific types when the distinction is meaningful.

\textbf{Guidelines:}
\begin{itemize}
\item Use specific types when the distinction is meaningful (e.g., ``athlete'' vs ``politician'' vs ``actor'')
\item But consolidate when entities serve the same role (all cities $\to$ ``location'', all companies $\to$ ``company'')
\item The SAME entity name can have different types in different contexts---read the full sentence to determine the correct interpretation
\item Before creating a new label, check if an existing one fits---use the same label for all entities of the same type
\item If the context doesn't provide enough information, use a broader category
\item Only assign ``NOISE'' if truly uninterpretable
\end{itemize}

\{existing\_block\}

\{tickets\_text\}

[Common output format appended here]
\end{quote}

\subsubsection{FewRel}

\paragraph{System prompt}
\begin{quote}\small
You are an expert at identifying semantic relations between entities. Your task is to classify the type of relation being expressed. The relationship direction and specificity matter. Read the context carefully to distinguish between similar relations.

\textbf{Few-Shot Examples:}
\begin{enumerate}
\item Input: ``John and Michael grew up together as brothers in Dublin. The relation between John and Michael.'' $\to$ Category: ``sibling''
\item Input: ``The Honda Civic is manufactured by Honda Motors in Japan. The relation between Honda Civic and Honda Motors.'' $\to$ Category: ``manufacturer''
\item Input: ``The British Museum holds the Rosetta Stone as a protected artifact. The relation between Rosetta Stone and British Museum.'' $\to$ Category: ``held by''
\item Input: ``Dr. Sarah Chen contributed research to the 2023 International Cancer Conference. The relation between Dr. Sarah Chen and 2023 International Cancer Conference.'' $\to$ Category: ``contributor to''
\end{enumerate}
\end{quote}

\paragraph{User prompt}
\begin{quote}\small
Here are \{n\_tickets\} entity relation descriptions. Identify the relation type for each.

\textbf{Guidelines for grouping:}
\begin{itemize}
\item Use clear relation names (e.g., ``field of work'', ``country of origin'', ``member of'', ``member of political party'')
\item Relations with the same semantic meaning belong together
\item Pay attention to directionality and specificity. For example:
  \begin{itemize}
  \item ``Albert Einstein'' and ``Physics'' with ``field of work'' relation is different from ``Albert Einstein'' and ``Germany'' with ``country of origin'' relation
  \item ``Marie Curie'' and ``Chemistry'' with ``field of work'' relation is different from ``Marie Curie'' and ``Poland'' with ``country of origin'' relation
  \end{itemize}
\item Some relations can be symmetric (e.g., ``sibling of'') vs directional (e.g., ``manufacturer of''). Direction matters for hierarchical relations (e.g., ``member of'' is different from ``has member'')
\item If a relation is unclear, assign ``NOISE''
\end{itemize}

\{existing\_block\}

\{tickets\_text\}

[Common output format appended here]
\end{quote}

\end{document}